\documentclass[conference]{IEEEtran}
\IEEEoverridecommandlockouts
\usepackage{cite} 
\usepackage{amsmath,amssymb,amsfonts}
\usepackage{algorithmic}
\usepackage{graphicx}
\usepackage{textcomp}
\usepackage{xcolor}
\usepackage{enumitem}
\usepackage{multirow}
\usepackage{booktabs}  
\usepackage{colortbl}  

\newcommand{\eg}{\textit{e}.\textit{g}.}
\usepackage{hyperref}

\def\BibTeX{{\rm B\kern-.05em{\sc i\kern-.025em b}\kern-.08em
    T\kern-.1667em\lower.7ex\hbox{E}\kern-.125emX}}
    
\begin{document}

\title{Sliced Maximal Information Coefficient: A Training-Free Approach for Image Quality Assessment Enhancement
\thanks{
This work was supported in part by the Natural Science Foundation of Guangdong Province (Grant 2023A1515011667), in part by Basic Research Foundation of Shenzhen (Grant JCYJ20210324093609026), and in part by Guangdong Basic and Applied Basic Research Foundation (Grant 2023B1515120020). Corresponding authors: Xuelin Shen and Baoliang Chen.}
}

\author{
    \IEEEauthorblockN{Kang Xiao$^{1,2}$, Xu Wang$^{2}$, Yulin He$^1$, Baoliang Chen$^{3\star}$, Xuelin Shen$^{1\star}$}
    \IEEEauthorblockA{$^1$ Guangdong Laboratory of Artificial Intelligence and Digital Economy (SZ), Shenzhen, China}
    \IEEEauthorblockA{$^2$ College of Computer Science and Software Engineering, Shenzhen University, Shenzhen, China}
    \IEEEauthorblockA{$^3$ School Of Computer Science, South China Normal University, Guangzhou, China}
    \IEEEauthorblockA{Email: shenxuelin@gml.ac.cn, blchen6-c@my.cityu.edu.hk}
}
\maketitle
\begin{abstract}
Full-reference image quality assessment (FR-IQA) models generally operate by measuring the visual differences between a degraded image and its reference. However, existing FR-IQA models including both the classical ones (\eg, PSNR and SSIM) and deep-learning based measures (\eg, LPIPS and DISTS) still exhibit limitations in capturing the full perception characteristics of the human visual system (HVS). In this paper, instead of designing a new FR-IQA measure, we aim to explore a generalized human visual attention estimation strategy to mimic the process of human quality rating and enhance existing IQA models.  In particular, we model human attention generation by measuring the statistical dependency between the degraded image and the reference image. The dependency is captured in a training-free manner by our proposed sliced maximal information coefficient and exhibits surprising generalization in different IQA measures. Experimental results verify the performance of existing IQA models can be consistently improved when our attention module is incorporated. The source code is available at https://github.com/KANGX99/SMIC.
\end{abstract}


\section{Introduction}\label{sec:intro}
Full-reference Image Quality Assessment (FR-IQA) aims to automatically generate a visual quality index for a distorted image by comparing it with the original reference image and has been widely employed in image restoration tasks, such as image compression,  denoising, deblurring, and super-resolution~\cite{ding2021comparison,chen2023gap,zhu2022enlightening}.
Many well-known FR-IQA models can be roughly classified as deterministic IQA, which calculate the difference between the reference image and distorted image based on explicit rules~\cite{laparra2016perceptual,fang2020blind,zhang2018unreasonable}.
To be specific, deterministic IQA methods provide the quality index by straightforwardly calculating the distance in certain spaces, such as Mean Squared Error (MSE) – the $\ell_2$ distance in the pixel domain, the Normalized Laplacian Pyramid Distance (NLPD) \cite{laparra2016perceptual} – the $\ell_2$ distance in the \textit{normalized Laplacian} domain, or LPIPS – $\ell_2$ distance in the deep feature space domain\cite{zhang2018unreasonable}. 
\begin{figure}[t]
      \centering
  \includegraphics[width=\linewidth]{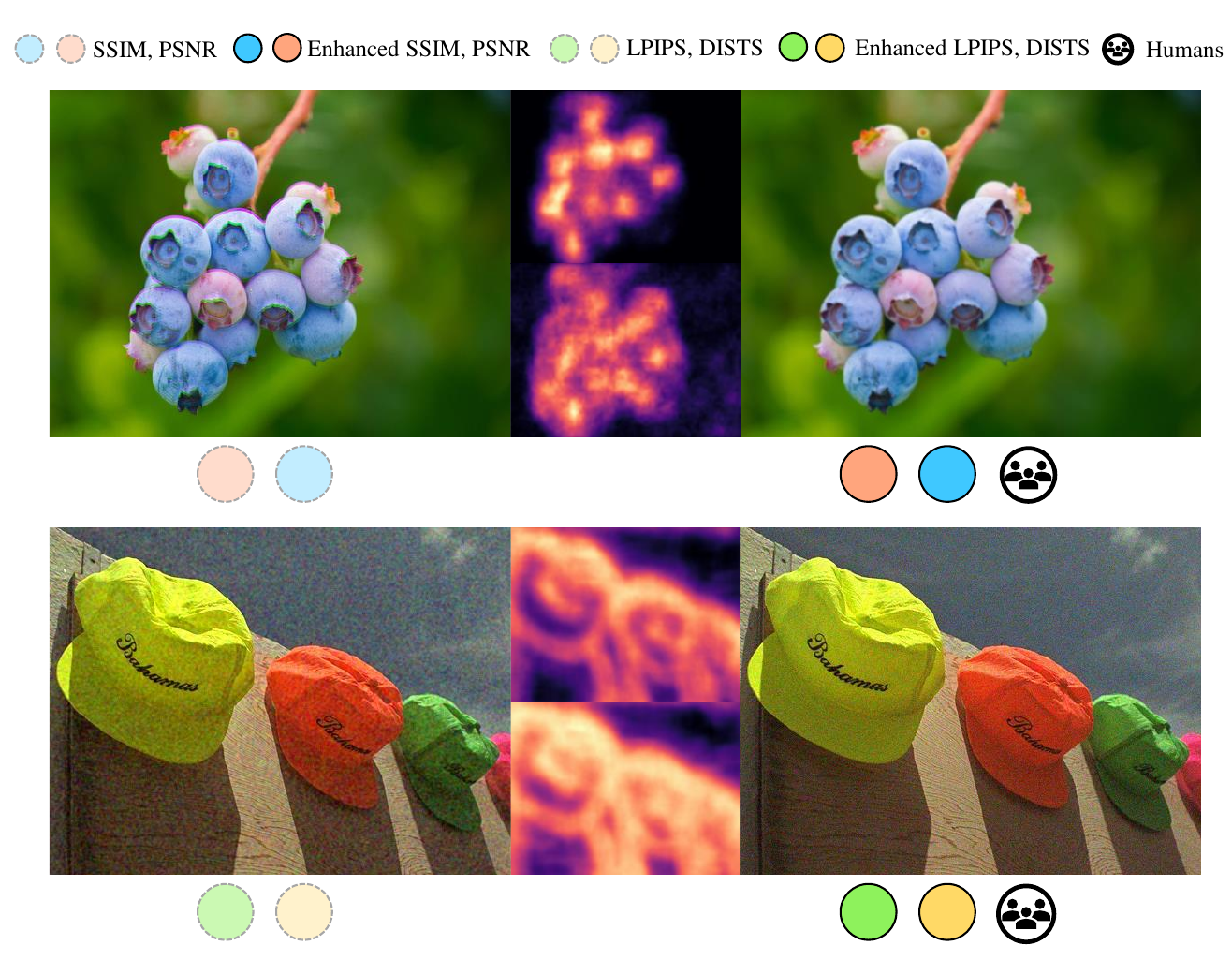}
  \caption{Existing FR-IQA models can be enhanced by our attention modeling.  Images in the first and third columns are different distorted images. Humans prefer the quality of the image in the third column, which is opposite to the prediction results of PSNR and SSIM (1st row), LPIPS and DISTS (2nd row). By incorporating the attention maps (shown in the second column), the enhanced FR-IQA models provide more consistent judgments with humans. We estimate human attention by the proposed Sliced Maximal Information Coefficient (SMIC) without any training process.}
  \label{fig:attentionmap}
\end{figure}
The explicit assessment roles result in a clear understanding and a computationally efficient evaluation process but lack proper consideration of human perception characteristics and are limited in handling complex distortions. 
Specifically, the methodology of deterministic IQA often exhibits high sensitivity to point-by-point deviations in texture regions, which are less aligned with human perceptions~\cite{ding2020image}.

Under these circumstances, statistical IQA, which involves comparing the statistical distributions of features between the reference image and the distorted image, has been widely studied.
Such as in~\cite{sheikh2006image}, Sheikh \textit{et al.} first modeled mutual information between the input image and images perceived by the Human Visual System (HVS) and employed mutual information divergence to provide a perceptual assessment index.
In~\cite{ding2020image}, Ding \textit{et al.} introduced structural and textural similarity in deep feature space to present a multi-scale, overcomplete understanding of image distortions.
In~\cite{liao2022deepwsd}, Liao \textit{et al.} modeled quality degradation from a statistical distribution perspective in deep feature space by employing the Wasserstein distance.

Meanwhile, some works attempt to further investigate the HVS's ``foveation characteristics", where only local image regions can be perceived in high resolution \cite{wang2004image}. These studies propose local-wise perceptual quality measurements instead of straightforwardly computing global assessments.
They share a two-stage methodology that first calculates the local-wise distortion measurement, which is then contributed to the final assessment via pooling.
Thus, plenty of efforts have been devoted to investigating the proper pooling strategies to mimic the spatial attention characteristics of HVS. For example,  an object-based pooling strategy, employed in \cite{gu2017blind, zhang2015som}, assigns more attention to object regions during pooling based on the straightforward observation that humans predominantly concentrate on such regions when viewing an image~\cite{judd2009learning}.
Meanwhile, other two-stage methods incorporate saliency detection technology to gain a powerful and comprehensive understanding of visually prominent and relevant regions within an image~\cite{lao2022attentions,gu2019blind,kim2017deep,bosse2017deep,zhang2014vsi,wang2010information,wang2006spatial}. However, the lack of generalizability and interpretability are still outstanding challenges of the data-driven attention generation models, resulting in their not utterly satisfied performance in actual implementation scenarios.

To construct robust computational two-stage FR-IQA models based on general biological principles, this paper reexamines the human attention estimation problem from the perspective of statistical dependence.
To be specific, an improved version of the Maximal Information Coefficient (MIC) \cite{reshef2011detecting} is proposed in this paper, named Sliced MIC (SMIC).
The proposed SMIC quantifies statistical dependency by calculating mutual information between the reference image and distorted image in deep feature space, contributing to robust and HVS-aligned attention generation, an intuitive example is shown in Fig.~\ref{fig:attentionmap}.
The resulting attention map (depicted in Fig. \ref{fig:attentionmap}) is subsequently employed for weighting the local distortion map which could be generated by both classic and deep-learning based IQA measures.
The main contributions of this paper can be summarized as
follows,
\begin{itemize}

    \item  A robust model for visual attention computation, supported by a theoretical framework, is established. The training-free advantage leads our attention estimation to be highly generalized to diverse FR-IQA frameworks.
    
    \item We propose the SMIC with both computational efficiency and ability regarding high-dimensional data analysis for mutual information computing.
   
    \item The experimental results underscore the effectiveness of the proposed attention estimation strategy for both learning-based and traditional IQA models across diverse distortion types, including GAN-based and super-resolution-based distortions.
    
\end{itemize}
\begin{figure}[t]
      \centering
  \includegraphics[width=\linewidth]{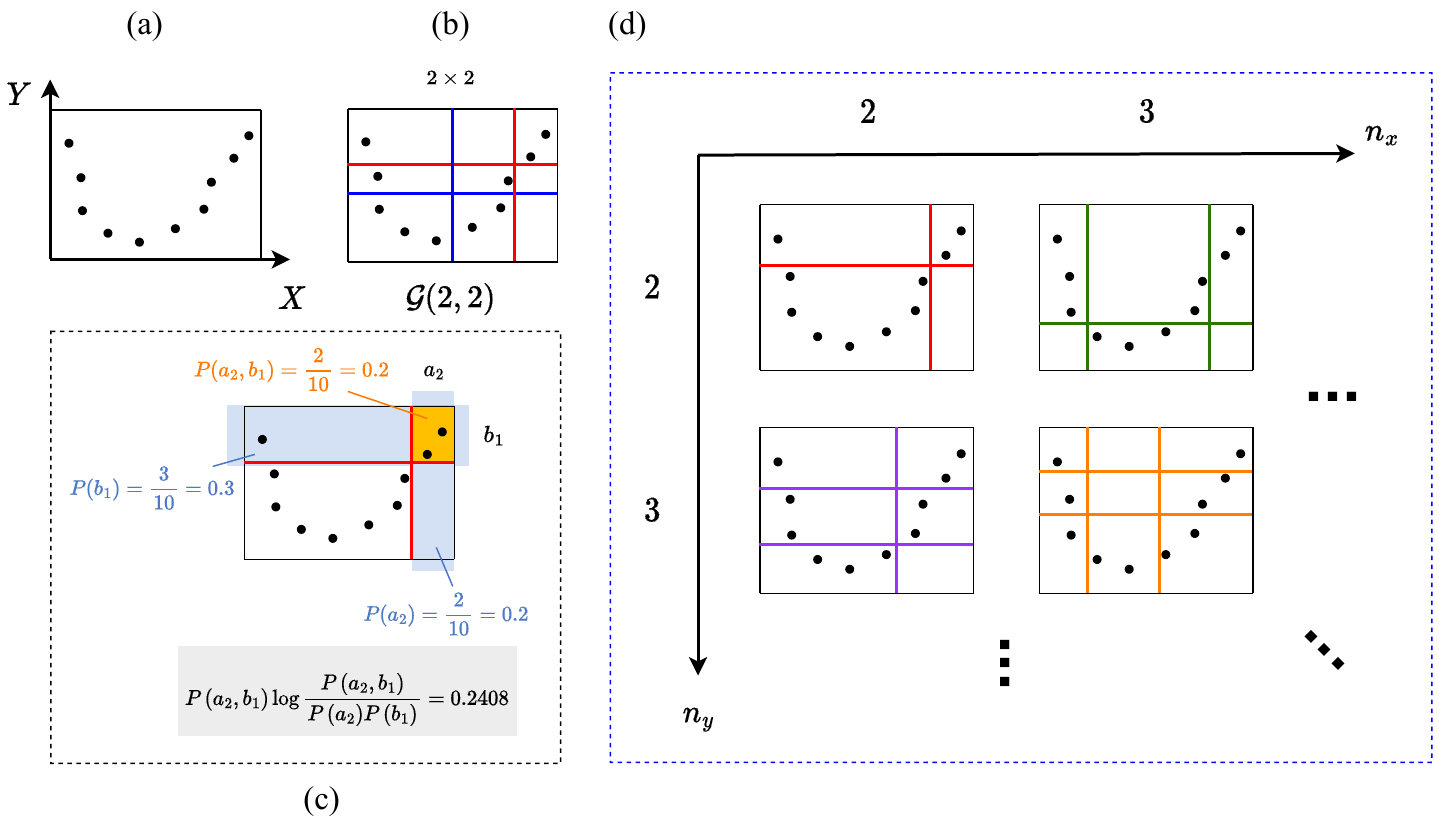}
  \caption{
  A toy example to illustrate the calculation of MIC between $X$ and $Y$.}
  \label{fig:mic}
\end{figure}
\section{Statistical Dependency-Guided Attention Weighting}

\subsection{Preliminary of Maximal Information Coefficient}
Given a paired variable ($X$, $Y$) and  their samples $\{x_1, x_2, ... , x_N\}$ and $\{y_1, y_2, ... , y_N\}$, the MIC aims to measure the statistic dependency between $X$ and $Y$. To calculate the MIC,  the samples of $X$ and $Y$ are first partitioned to the $x$-axis and $y$-axis in order, forming a 2D scatterplot.  The MIC is then defined by,
\begin{equation}
    \textrm{MIC}(X,Y)=\operatorname*{max}_{n_x,n_y: n_x*n_y<N^{0.5}}\big\{\frac{\operatorname*{max}_{G\in \mathcal{G}(n_x,n_y)}I_G(X,Y)}{\log\operatorname*{min}\{n_x,n_y\}}\big\},
    \label{eqn:mic}
\end{equation}
where $N$ is the number of samples, and $n_x$ and $n_y$ are the number of grid bins on the $x$-axis and $y$-axis, respectively. $\mathcal{G}(n_x,n_y)$ is the set of all possible grids of the scatterplot with the size $n_x * n_y$.  $I_G(X,Y)$ denotes the mutual information under a specific grid $G$. 
The $\log\operatorname*{min}\{n_x,n_y\}$ is a normalization term to ensure MIC in the range [0,1].
In Fig. \ref{fig:mic}, we present a toy example to illustrate the calculation details. In particular, we first place all the paired samples of $X$ and $Y$ into a scatterplot (see Fig. \ref{fig:mic}(a)). Then we grid the scatterplot by  $n_x$-by-$n_y$ cells. Herein, numerous grid schemes can be adopted when  $n_x>1$ and $n_y>1$.  In Fig. \ref{fig:mic}(b), we show two possible grids (in red and blue) when $n_x=n_y=2$. For a possible grid $G$, we measure the mutual information $I_G(X,Y)$ by the following formula,
\begin{equation}
    I_G(X,Y)=\sum_{u=1}^{n_x} \sum_{v=1}^{n_y} P(a_u, b_v) \log \frac{P(a_u, b_v)}{P(a_u) P(b_v)},
    \label{eqn:mi}
\end{equation}
where $a_u$ denotes the points in the $u$-th region along the horizontal direction formed by vertical grid lines, and $b_v$ represents the points in the $v$-th region along the vertical direction created by horizontal grid lines (see Fig. \ref{fig:mic}(c)).
Different grids will result in multiple mutual information values. We select and normalize the maximum one as the estimated mutual information under the grid set  $\mathcal{G}(2,2)$. Finally, we adjust $n_x$ and $n_y$ under $n_x*n_y < N^{0.5}$ and the MIC is the maximum value under all the grid sets (see Fig. \ref{fig:mic}(d)).

\begin{figure*}[htp]
      \centering
  \includegraphics[width=0.95\linewidth]{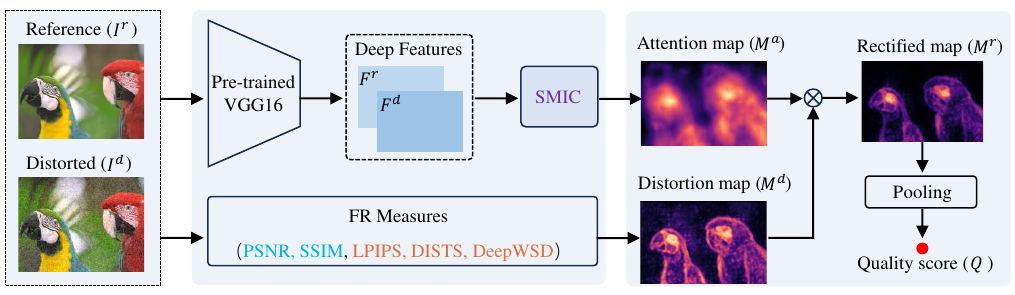}
  \caption{The framework of our enhancement strategy for existing FR-IQA models. }
\label{fig:archi}
\end{figure*}
Compared to other methods, MIC better captures diverse dependencies between variables $X$ and $Y$ \cite{albanese2013minerva}. However, exact MIC computation is challenging due to the need to consider all possible grids. Our approach employs a heuristic MIC approximation algorithm \cite{reshef2011detecting}, enabling more efficient measurement without much accuracy sacrifice.

\subsection{Enhance existing FR-IQA models by SMIC}
Existing FR-IQA models especially the deep-learning based models usually estimate the image quality in a two-stage manner, including 1) the distortion map generation by comparing the reference image and distorted image, and 2) the quality score estimation by aggregating the distortion map. However, naïve aggregation strategies (average pooling or std pooling)  usually lead the quality predictions not always consistent with human opinions. Inspired by the human quality rating process, we propose an attention-guided aggregation module for existing FR-IQA model enhancement. We model the human attention generation based upon the MIC under a mild assumption that the higher the dependence between the regions in the reference image and distorted image, the attention would be lower. Our framework is illustrated in Fig.~\ref{fig:archi}, which mainly contains three steps: 1) Distortion map generation by existing FR-IQA models. 2) Attention map generation by the proposed SMIC. 3) Quality score acquisition. We detail the three steps in the following sub-paragraphs.

\subsubsection{Distortion Map Generation by Existing FR-IQA Models}
For traditional FR-IQA models, we select two classical measures, PSNR and SSIM \cite{wang2004image} for the performance validation. In particular, we obtain the distortion map ($M^d$) by comparing the difference between the reference image ($I^r$) and the distorted image ($I^d$) in the pixel domain. 

For deep-learning based FR-IQA models, the popular VGG \cite{simonyan2014very} based measures including the LPIPS \cite{zhang2018unreasonable}, DISTS \cite{ding2020image} and DeepWSD \cite{liao2022deepwsd} are selected. To obtain the distortion map, we measure the feature difference in a patch manner with their default comparison metrics. Supposing the FR-IQA model is LPIPS and the features of $I^r$ and $I^d$ extracted from  the $s$-th stage of the VGG16 network are $F^r_{s}$ and $F^d_{s}$,  the distortion map $M_s^d$ can be obtained by,
\begin{equation}
    M_{s,p}^d(F^r_{s,p},F^d_{s,p}) = \frac{1}{H_sW_s}\left \| F^r_{s,p} - F^d_{s,p} \right \|_2^2
    \label{eqn:distmap}
\end{equation}
where $p$ is the spatial index and $F^r_{s, p}$ and $F^d_{s, p}$ are the $p$-th patch in $F^r_{s}$ and $F^d_{s}$. Herein, $F^r_{s,p}$ and $F^d_{s,p} \in \mathbb{R}^{H_s\times W_s\times C_s}$ where  $H_s$ and $W_s$ are the height and width of each patch. The $C_s$ represents the channel number of each feature. The  $M_{s,p}^d$ is the value of $M_{s}^d$ at the location $p$.

\begin{table*}[h]
\caption{Performance comparison on six standard IQA databases. Larger SRCC and PLCC values indicate that the IQA model is more consistent with the HVS perception. The best, the second-best, and the third-best results are highlighted in boldface\&underlined, boldface, and underlined, respectively. Additionally, `$\textcolor{blue}{\uparrow}$' and `\textcolor{gray}{-}' represent performance improvement, unchanged respectively.}
\label{tab:mainresult}
\centering
\resizebox{\textwidth}{!}{ 
\begin{tabular}{lcccccccccccc}
\toprule
\multirow{2}{*}{Method}&\multicolumn{2}{c}{LIVE \cite{sheikh2005live}} & \multicolumn{2}{c}{CSIQ \cite{larson2010most}} & \multicolumn{2}{c}{TID2013 \cite{ponomarenko2015image}} & \multicolumn{2}{c}{KADID-10k \cite{lin2019kadid}} & \multicolumn{2}{c}{PIPAL \cite{jinjin2020pipal}} & \multicolumn{2}{c}{QADS \cite{zhou2019visual}} \\
\cmidrule(l){2-3}\cmidrule(l){4-5}\cmidrule(l){6-7}\cmidrule(l){8-9}\cmidrule(l){10-11}\cmidrule(l){12-13}

&SRCC & PLCC & SRCC & PLCC & SRCC & PLCC  &SRCC & PLCC & SRCC & PLCC & SRCC & PLCC \\
\midrule
MS-SSIM \cite{wang2003multiscale}&0.951&0.943&0.917&0.902&0.781&0.827&0.803&0.801&0.559&0.585&0.717&0.721\\
FSIM \cite{zhang2011fsim}&\textbf{0.961}&0.949&0.931&0.919&0.851&\underline{0.877}&0.854&0.851&0.590&0.613&0.687&0.689\\
VIF \cite{sheikh2006image}&\underline{\textbf{0.972}}&\underline{\textbf{0.972}}&0.919&0.926&0.677&0.772&0.791&0.793&0.539&0.560&0.815&0.821 \\
NLPD \cite{laparra2016perceptual}&0.945&\underline{0.957}&0.937&0.927&0.800&0.843&0.844&0.813&0.483&0.507&0.591&0.605 \\
MAD \cite{larson2010most}&0.957&0.949&0.947&0.950&0.781&0.827&0.797&0.823&0.559&0.611&0.723&0.731 \\
PieAPP \cite{prashnani2018pieapp}&0.933&0.945&0.897&0.880&0.848&0.835&0.786&0.789&\underline{\textbf{0.706}}&\textbf{0.709}&\underline{\textbf{0.861}}&\underline{\textbf{0.863}} \\
\midrule
PSNR&0.923&0.920&0.849&0.844&0.700&0.675&0.604&0.612&0.501&0.520&0.573&0.587 \\
SSIM \cite{wang2004image}&0.930&0.928&0.866&0.853&0.716&0.742&0.717&0.716&0.557&0.585&0.713&0.719 \\
LPIPS \cite{zhang2018unreasonable}&0.932&0.935&0.883&0.906&0.670&0.759&0.720&0.700&0.573&0.611&0.671&0.674 \\
DISTS \cite{ding2020image}&0.954&0.954&0.939&0.941&0.830&0.856&0.887&0.886&0.623&0.644&0.809&0.808 \\
DeepWSD \cite{liao2022deepwsd}&0.952&0.949&\textbf{0.963}&\textbf{0.953}&\textbf{0.874}&0.876&0.888&0.888&0.514&0.517&0.762&0.770\\
\midrule
PSNR \textit{w/} SMIC&0.945  ($\textcolor{blue}{\uparrow2.2\%}$)&0.941  ($\textcolor{blue}{\uparrow2.1\%}$)&0.917 ($\textcolor{blue}{\uparrow6.8\%}$)&0.881 ($\textcolor{blue}{\uparrow3.7\%}$)&0.826 ($\textcolor{blue}{\uparrow12.6\%}$)&0.797 ($\textcolor{blue}{\uparrow12.2\%}$)&0.763 ($\textcolor{blue}{\uparrow15.9\%}$)&0.749 ($\textcolor{blue}{\uparrow13.7\%}$)&0.566 ($\textcolor{blue}{\uparrow6.5\%}$)&0.583 ($\textcolor{blue}{\uparrow6.3\%}$)&0.698 ($\textcolor{blue}{\uparrow12.5\%}$)&0.704 ($\textcolor{blue}{\uparrow11.7\%}$) \\
SSIM \textit{w/} SMIC&0.947 ($\textcolor{blue}{\uparrow1.7\%}$)&0.943 ($\textcolor{blue}{\uparrow1.5\%}$)&0.929
  ($\textcolor{blue}{\uparrow6.3\%}$)&0.916 ($\textcolor{blue}{\uparrow6.3\%}$)&0.801 ($\textcolor{blue}{\uparrow8.5\%}$)&0.762 ($\textcolor{blue}{\uparrow2.0\%}$)&0.815 ($\textcolor{blue}{\uparrow9.8\%}$)&0.812 ($\textcolor{blue}{\uparrow9.6\%}$)&0.605 ($\textcolor{blue}{\uparrow4.8\%}$)&0.632 ($\textcolor{blue}{\uparrow4.7\%}$)&0.760 ($\textcolor{blue}{\uparrow4.7\%}$)&0.768 ($\textcolor{blue}{\uparrow4.9\%}$) \\
LPIPS \textit{w/} SMIC&0.959 ($\textcolor{blue}{\uparrow2.7\%}$)&0.946 ($\textcolor{blue}{\uparrow1.1\%}$)&\underline{0.962} ($\textcolor{blue}{\uparrow7.9\%}$)&\underline{0.951} ($\textcolor{blue}{\uparrow4.5\%}$)&\underline{0.867} ($\textcolor{blue}{\uparrow19.7\%}$)&\textbf{0.891} ($\textcolor{blue}{\uparrow13.2\%}$)&\underline{\textbf{0.911}} ($\textcolor{blue}{\uparrow19.1\%}$)&\underline{\textbf{0.907}} ($\textcolor{blue}{\uparrow20.7\%}$)&\underline{0.657} ($\textcolor{blue}{\uparrow8.4\%}$)&\underline{0.665} ($\textcolor{blue}{\uparrow5.4\%}$)&\textbf{0.839} ($\textcolor{blue}{\uparrow16.8\%}$)&\textbf{0.842} ($\textcolor{blue}{\uparrow16.8\%}$) \\
DISTS \textit{w/} SMIC&\underline{0.960} ($\textcolor{blue}{\uparrow0.6\%}$)&0.954 (\textcolor{gray}{- 0.0\%})&0.956 ($\textcolor{blue}{\uparrow1.7\%}$)&0.947 ($\textcolor{blue}{\uparrow0.6\%}$)&0.851 ($\textcolor{blue}{\uparrow2.1\%}$)&0.871 ($\textcolor{blue}{\uparrow1.5\%}$)&\underline{0.890} ($\textcolor{blue}{\uparrow0.3\%}$)&\underline{0.889} ($\textcolor{blue}{\uparrow0.3\%}$)&\textbf{0.673} ($\textcolor{blue}{\uparrow5.0\%}$)&\underline{\textbf{0.709}} ($\textcolor{blue}{\uparrow6.5\%}$)&\underline{0.835} ($\textcolor{blue}{\uparrow2.6\%}$)&\underline{0.839} ($\textcolor{blue}{\uparrow3.1\%}$) \\
DeepWSD \textit{w/} SMIC&0.959 ($\textcolor{blue}{\uparrow0.7\%}$)&\textbf{0.957} ($\textcolor{blue}{\uparrow0.8\%}$)&\underline{\textbf{0.967}} ($\textcolor{blue}{\uparrow0.4\%}$)&\underline{\textbf{0.965}} ($\textcolor{blue}{\uparrow1.2\%}$)&\underline{\textbf{0.882}} ($\textcolor{blue}{\uparrow0.8\%}$)&\underline{\textbf{0.903}} ($\textcolor{blue}{\uparrow2.7\%}$)&\textbf{0.906} ($\textcolor{blue}{\uparrow1.8\%}$)&\textbf{0.906} ($\textcolor{blue}{\uparrow1.8\%}$)&0.608 ($\textcolor{blue}{\uparrow9.4\%}$)&0.629 ($\textcolor{blue}{\uparrow11.2\%}$)&0.814 ($\textcolor{blue}{\uparrow5.2\%}$)&0.820 ($\textcolor{blue}{\uparrow5.0\%}$)\\
\bottomrule
\end{tabular}
}
\end{table*}
\subsubsection{Attention Map Generation by the Proposed SMIC}\label{subsubsec:limitation}
In this step, instead of estimating the statistical dependency between $I^r$ and $I^d$ in the pixel domain by MIC directly, we propose the SMIC to capture their dependency in the deep-feature domain. The reasons lie in that: 1) Compared with the RGB values, the deep-features extracted by the pre-trained VGG network are more quality-aware \cite{zhang2018unreasonable, chen2021learning}, leading the dependency measured in the deep-feature space to be more consistent with HVS. 2) As we depicted in Sec. II.A, the MIC is only applicable for one-dimensional random variables. However, the deep features are usually with high dimensions, leading the original MIC to be less effective. Motivated by the sliced mutual information (SMI) \cite{goldfeld2021sliced}, the SMIC is proposed for the dependency estimation in the deep-feature domain. The following are the details.

\noindent \textbf{Project the image into the deep-feature domain.}
To obtain quality-aware features, we still adopt the popular pre-trained VGG16 network as the feature extractor. In particular, we extract muti-scale deep features at different stages of the VGG16 network.

\noindent \textbf{Measure the feature dependency  by SMIC.}
Following the idea of SMI \cite{goldfeld2021sliced}, we adopt random projections to project the high-dimensional features into different one-dimensional variables. The SMIC thus can be estimated by averaging the MIC values formed by those one-dimensional variables, which can be depicted as follows,
\begin{figure}[t]
      \centering
  \includegraphics[width=\linewidth]{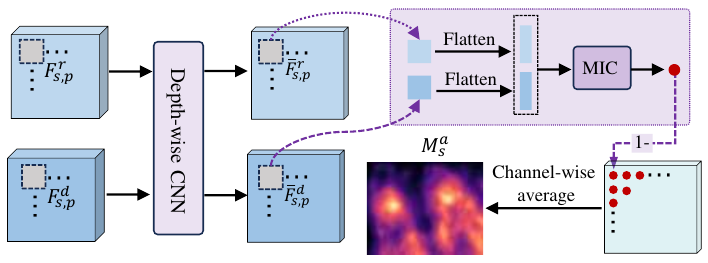}
  \caption{Illustration of the attention map generation by our proposed SMIC.}
  \label{fig:smic}
\end{figure}
\begin{equation}  
\textrm{SMIC}(X,Y) ={\frac{1}{S_{C_s-1}^2}}\oint_{\mathbb{S}^{C_s-1}}\oint_{\mathbb{S}^{C_s-1}} \textrm{MIC}(\Theta^{\top}X, \Phi^{\top}Y)\mathrm{d}\theta\mathrm{d}\phi,
\label{eqn:smic}
\end{equation}
where $\mathbb{S}^{C_s-1}$ is the $C_s$-dimensional unit sphere, and its surface area is $S_{C_s-1} = 2\pi^{C_s/2}/\Gamma(C_s/2)$, with $\Gamma$ as the
gamma function.  $\Theta \sim \operatorname{Unif}\left(\mathbb{S}^{C_s-1}\right)$ and $\Phi \sim \operatorname{Unif}\left(\mathbb{S}^{C_s-1}\right)$ are independent of each other and of $(X,Y)$. The SMIC inherits key properties of SMI, such as discrimination between dependence and independence, chain rule, etc.
In our method, we treat the $F^r_{s,p}$ and $F^d_{s,p}$ as samples drawn  from the high-dimensional variables $X\in \mathbb{R}^{C_s}$ and $Y\in \mathbb{R}^{C_s}$, then their SMIC can be efficiently approximated by the Monte Carlo sampling,
\begin{equation}
    \textrm{SMIC}\left(F^d_{s,p}, F^r_{s,p} \right)\approx \frac{1}{K} \sum_{i=1}^{K} \textrm{{MIC}}\left(\Theta_i^{\top}F^r_{s,p}, \Phi_i^{\top}F^d_{s,p}\right),
    \label{eqn:smic_e}
\end{equation}
where $K$ is the sampling times. As shown in Fig.~\ref{fig:smic}, we implement the $K$ projections by a depth-wise convolutional layer with the output channel number set as $K$. Herein, the convolution weights are sampled from a pre-set Gaussian distribution without any training process. Once the SMIC of all the  patches is calculated, we could obtain the SMIC map  $M_{s}$ of  ${F}^r_{s}$ and ${F}^d_{s}$ and the attention map is measured by, 
\begin{equation}
M_{s}^a = 1-M_{s},
\end{equation}
with the assumption that a higher feature dependency results in lower attention during human quality rating.

\subsection{Quality Score Acquisition}
For traditional FR-IQA measures, including the PSNR and SSIM \cite{wang2004image}, the final quality score $Q(I^r,I^d)$ can be obtained by,

\begin{equation}
    Q(I^r,I^d) = f(\frac{\sum_{s=m}^{n}\mathcal{R}_s(M_{s}^a)}{m-n+1} {\otimes} M^{d}),
    \label{eqn:tra}
\end{equation}
where $m$ and $n$ are the first and last stage indexes we selected for attention map generation.  $\mathcal{R}_s(\cdot)$ means a resize operation to align the dimension between $M_{s}^a$ and $M^d$. The ``$\otimes$" means element-wise multiplication. $f(\cdot)$ represents the average operation.

For VGG-based metrics, including the LPIPS \cite{zhang2018unreasonable}, DISTS \cite{ding2020image} and DeepWSD \cite{liao2022deepwsd}, we estimate the quality score $Q(I^r,I^d)$ by,
\begin{equation}
    Q(I^r,I^d) = \frac{\sum_{s=m}^{n}f\left (M_{s}^a{\otimes} M^{d}_s\right )}{m-n+1}  + \sum_{s\notin \lbrace m,m+1,...,n\rbrace}\mathcal{D}(F^r_s,F^d_s) ,
    \label{eqn:vgg}
\end{equation}
where $\mathcal{D}(\cdot)$ represents the specific feature difference measure proposed in the corresponding FR-IQA model. 


\begin{figure*}[htbp]
      \centering
  \includegraphics[width=\linewidth]{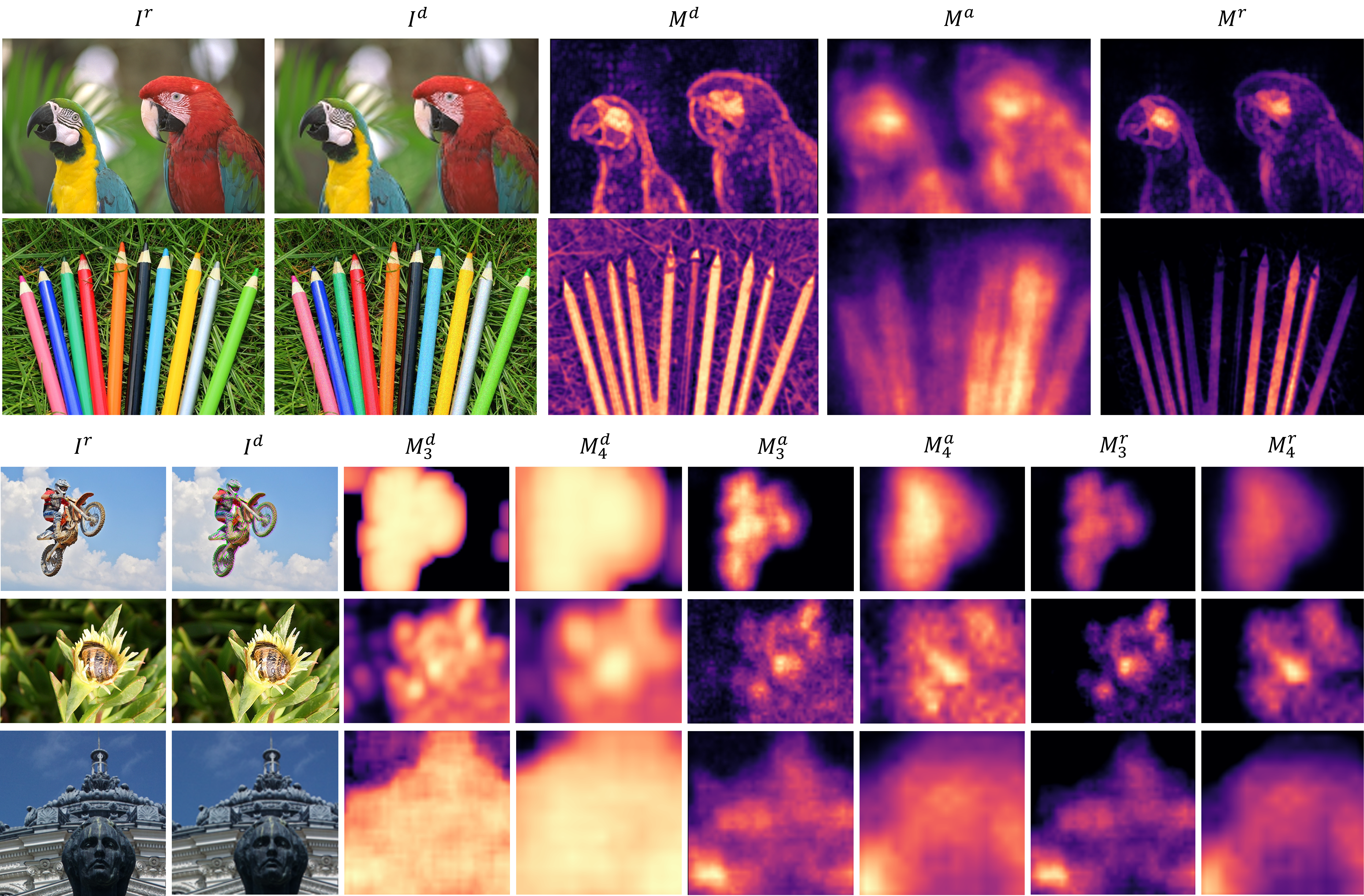}
  \caption{Visualisation of the generated attention maps. $I^r$ and $I^d$ are the reference and distorted images. $M^d$ represents the distortion maps generated by PSNR (1st row) and SSIM (2nd row).  $M^d_s$ (s = 3,4) are the distortion maps measured  by LPIPS (3rd row), DISTS (4th row), and DeepWSD (5th row) at the $s$-th stage of the VGG16 network.  $M^a$ and $M^a_s$ are the estimated attention maps.  $M^r$ and $M^r_s$ are the rectified distortion map with $M^r = M^d \otimes M^a$ and $M^r_s = M^d_s \otimes M^a_s$. 
  }
  \label{fig:errormap}
\end{figure*}
\section{Experiments}
\subsection{Implementation Details}
During the distortion map generation, we set the patch size as $7\times 7$, and the patch stride is set as  $1$ and $7$ for the traditional and deep-learning based IQA models, respectively. We set $K$ as $32$ in Eqn. (\ref{eqn:smic_e}) and  $m=3$ and $n=4$ in Eqn. (\ref{eqn:tra}).  $H_s = W_s =7$ in Eqn. (\ref{eqn:distmap}). The bilinear interpolation is adopted for the resize operation $\mathcal{R}_s(\cdot)$ in Eqn. (\ref{eqn:tra}).

\subsection{Benchmark}
The proposed SMIC-based method is evaluated based on six IQA datasets: TID2013 \cite{ponomarenko2015image}, LIVE \cite{sheikh2005live}, CSIQ \cite{larson2010most}, KADID-10k \cite{lin2019kadid}, PIPAL \cite{jinjin2020pipal}, and QADS \cite{zhou2019visual}.
Specifically, the QADS is specifically designed for super-resolution-oriented distortions. PIPAL, the largest human-rated IQA dataset, notably includes 19 additional GAN-based distortion types, posing a significant challenge to existing FR-IQA methods.
SRCC and PLCC are employed as evaluation criteria.
Furthermore, a five-parameter nonlinear logistic function is employed to align the predicted scores and MOSs.

The SMIC-based weighting scheme is integrated into five FR-IQA methods, including classical metrics (PSNR and SSIM \cite{wang2004image}) and learning-based methods (LPIPS \cite{zhang2018unreasonable}, DISTS \cite{ding2020image}, DeepWSD \cite{liao2022deepwsd}).
%
Improved versions of these metrics are compared with six state-of-the-art methods: MS-SSIM \cite{wang2003multiscale}, FSIM \cite{zhang2011fsim}, VIF \cite{sheikh2006image}, NLPD \cite{laparra2016perceptual}, MAD \cite{larson2010most}, PieAPP \cite{prashnani2018pieapp}.

\subsection{Experimental Results}
\noindent \textbf{Quantitative Evaluation.} The comparison results are shown in Table \ref{tab:mainresult}, where `\textit{w/}' denotes the incorporation of our proposed scheme.
Quite encouraging results have been demonstrated that the proposed weighting scheme brings an overall improvement to the employed traditional and learning-based FR-IQA models across six datasets.
The demonstrated effectiveness across synthetic, GAN, and super-resolution-oriented distortions underscores the strong robustness of the proposed weighting scheme, owing to the trustworthy and interpretable framework employed.
Moreover, incorporating the SMIC-based weighting enhances even basic metrics like PSNR to outperform state-of-the-art models, underlining the necessity of HVS characteristics in IQA modeling.
%
%

\noindent \textbf{Qualitative Evaluation.} To gain intuition into the attention modeling capacity of the proposed scheme, an example is illustrated in Fig. \ref{fig:errormap}.
From the Figure, we can observe the attention maps generated with our method successfully model visual attention, where areas with higher attention coincide with regions where distortion is more noticeable to humans, such as the foreground in the first row and object regions in the second and third rows.
In contrast, compared with the local distortion map generated by existing IQA methods ($M^d$ and $M^d_s$ ), the overemphasis on regions that are less noticeable to humans can be easily observed, such as the blurred background in the first and fourth rows, resulting in their unsatisfactory performance.
Moreover, these inconsistencies are successfully rectified by employing the SMIC-based weighting scheme ($M^r$ and $M^r_s$), leading to improved performance.

\noindent \textbf{Effectiveness of SMIC.}
\begin{table}
\caption{Performance comparison between MIC and SMIC for human attention modeling.}
\label{tab:smic_effect}
\centering
\resizebox{\linewidth}{!}{ 
\begin{tabular}{lcccccccc}
\toprule
\multirow{2}{*}{Method}&\multicolumn{2}{c}{LIVE \cite{sheikh2005live}} & \multicolumn{2}{c}{KADID-10k \cite{lin2019kadid}} & \multicolumn{2}{c}{PIPAL \cite{jinjin2020pipal}} & \multicolumn{2}{c}{QADS \cite{zhou2019visual}} \\
\cmidrule(l){2-3}\cmidrule(l){4-5}\cmidrule(l){6-7}\cmidrule(l){8-9}

&SRCC & PLCC & SRCC & PLCC & SRCC & PLCC  &SRCC & PLCC \\
\midrule
PSNR \textit{w/} MIC&0.930&0.919&0.632&0.637&0.518&0.538&0.594&0.607\\
SSIM \textit{w/} MIC&0.936&0.931&0.732&0.732&0.570&0.598&0.726&0.730\\
LPIPS \textit{w/} MIC&0.955&0.945&0.899&0.898&\underline{0.645}&0.660&\underline{0.823}&\underline{0.830}\\
DISTS \textit{w/} MIC&0.947&0.947&0.819&0.821&0.621&\underline{0.662}&0.818&0.815\\
DeepWSD \textit{w/} MIC&0.958&\textbf{0.956}&\underline{0.902}&\underline{0.902}&0.601&0.621&0.811&0.813\\
\midrule
PSNR \textit{w/} SMIC&0.945&0.941&0.763&0.749&0.566&0.583&0.698&0.704
\\
SSIM \textit{w/} SMIC&0.947&0.943&0.815&0.812&0.605&0.632&0.760&0.768 \\
LPIPS \textit{w/} SMIC&\underline{0.959}&0.946&\underline{\textbf{0.911}}&\underline{\textbf{0.907}}&\textbf{0.657}&\textbf{0.665}&\underline{\textbf{0.839}}&\underline{\textbf{0.842}}\\
DISTS \textit{w/} SMIC&\underline{\textbf{0.960}}&\underline{0.954}&0.890&0.889&\underline{\textbf{0.673}}&\underline{\textbf{0.709}}&\textbf{0.835}&\textbf{0.839}\\
DeepWSD \textit{w/} SMIC&\textbf{0.959}&\underline{\textbf{0.957}}&\textbf{0.906}&\textbf{0.906}&0.608&0.629&0.814&0.820\\
\bottomrule
\end{tabular}
}
\label{tbl:smic_eff}
\end{table}
In our method, we propose SMIC to alleviate the limitation of MIC in capturing the statistical dependency of high-dimensional variables. To verify the effectiveness of SMIC, we use MIC to measure the feature dependency directly and compare the results with our original one in Table \ref{tbl:smic_eff}. From the table, we can observe that the performance of SMIC-based measures outperforms the MIC-based ones in terms of both PLCC and SRCC. The phenomenon can be observed consistently in different IQA datasets, revealing the high superiority of our proposed SMIC. 

\section{Conclusions}
In this paper, we propose a training-free attention modeling scheme for enhancing existing FR-IQA models.
To be specific, we optimize the Maximal Information Coefficient (MIC) to enable the statistical dependency estimation in high dimensional deep feature space, contributing to human attention modeling.
By incorporating trustworthy and interpretable attention modeling into local-wise distortion pooling, existing FR-IQA models are better aligned with HVS characteristics, resulting in favorable performance improvement.
Extensive experimental results across multiple FR-IQA models and test datasets highly demonstrate the effectiveness of the proposed SMIC-based weighting scheme.
\bibliographystyle{IEEEtran}
\bibliography{main3}

\end{document}